\def\year{2019}\relax
\documentclass[letterpaper]{article} 
\usepackage{aaai19}  
\usepackage{times}  
\usepackage{helvet}  
\usepackage{courier}  
\usepackage{url}  
\usepackage{graphicx}  
\usepackage{multirow}
\usepackage{algorithm}
\usepackage{algorithmic}
\usepackage{amsmath,bm,amssymb,amsthm}
\usepackage{amsfonts,amstext}
\usepackage{booktabs}
\usepackage{subfig}

\graphicspath{{./figures/}}
\newtheorem{definition}{Definition}
\DeclareMathOperator*{\argmin}{arg\,min}

\title{Skeleton-based Activity Recognition\\ with Local Order Preserving Match of Linear Patches}

\author{
Yaqiang Yao, Yan Liu, Huanhuan Chen, \\
School of Computer Science and Technology, \\
University of Science and Technology of China, Hefei, China\\
yaoyaq@mail.ustc.edu.cn, ready@mail.ustc.edu.cn, hchen@ustc.edu.cn}

\begin{document}

\nocopyright
\maketitle

\begin{abstract}
Human activity recognition has drawn considerable attention recently in the field of computer vision due to the development of commodity depth cameras, by which the human activity is represented as a sequence of 3D skeleton postures. Assuming human body 3D joint locations of an activity lie on a manifold, the problem of recognizing human activity is formulated as the computation of activity manifold-manifold distance (AMMD). In this paper, we first design an efficient division method to decompose a manifold into ordered continuous maximal linear patches (CMLPs) that denote meaningful action snippets of the action sequence. Then the CMLP is represented by its position (average value of points) and the first principal component, which specify the major posture and main evolving direction of an action snippet, respectively. Finally, we compute the distance between CMLPs by taking both the posture and direction into consideration. Based on these preparations, an intuitive distance measure that preserves the local order of action snippets is proposed to compute AMMD. The performance on two benchmark datasets demonstrates the effectiveness of the proposed approach.
\end{abstract}

\section{Introduction}
\label{sec:introduction}
In computer vision and machine learning communities, human activity recognition has become one of the most appealing studies \cite{vrigkas2015review,xu2017learning} for its wide applications. Previous RGB-based work focused on extracting local space-time features from 2D images. Recently, with the introduction of real-time depth cameras and corresponding human skeleton extraction methods \cite{shotton2013real}, the studies of activity recognition have been greatly promoted in terms of depth maps-based methods \cite{rahmani2016histogram} and skeleton-based methods \cite{wang2014learning}. In particular, \cite{yao2011does} verified that skeleton data alone can outperform other low-level image features for human activity recognition. The main reason is that the 3D skeleton poses are invariant to the viewpoint and appearance, such that activities vary less from actor to actor. Several specially designed descriptors such as HOJ3D \cite{xia2012view}, Cov3DJ \cite{hussein2013human}, and HOD \cite{gowayyed2013histogram} employed the conclusion and achieve decent performance.

\subsection{Related Work}
Different from human posture recognition, the temporal relation between adjacent frames poses a challenge to the activity recognition task. Human body represented with 3D skeleton can be viewed as an articulated system in which rigid segments are connected with several joints. In this way, we can treat a human activity as an evolution of the spatial configuration of these segments. Based on this perspective, \cite{gong2014structured} addressed the human activity recognition as the problem of structured time series classification. From the view of temporal dynamics modeling, existing methods for human activity recognition fall into two categories: state space models and recurrent neural networks (RNNs). State space models, including linear dynamic system \cite{chaudhry2013bio}, hidden Markov model \cite{lv2006recognition,wu2014leveraging} and conditional restricted Boltzmann machine \cite{taylor2007modeling}, treated the action sequence as an observed output produced by a Markov process, whose hidden states are used to model the dynamic patterns. In contrast, RNNs utilize their internal state (memory) instead of dynamic patterns to process an action sequence of inputs \cite{du2015hierarchical,zhu2016co,song2017end}. However, \cite{elgammal2004inferring} showed that the geometric structure of activity would not be preserved in temporal relation and proposed to learn the representation of activity with manifold embedding. In particular, the authors nonlinearly embedded activity manifolds into a low dimensional space with LLE and found that the temporal relation of the input sequence was preserved in the obtained embedding to some extent.

Manifold based representation and related algorithms have attracted much attention in image and video analysis. In consideration of temporal dimension, \cite{wang2007learning} exploited locality preserving projections to project a given sequence of moving silhouettes associated with an action video into a low-dimensional space. Modeling each image set with a manifold, \cite{wang2012manifold} formulated the image sets classification for face recognition as a problem of calculating the manifold-manifold distance (MMD). The authors extracted maximal linear patches (MLPs) to form nonlinear manifold and integrated the distances between pairs of MLPs to compute MMD. Similar to image sets that each set is composed of images from the same person but covering variations, human body 3D joint locations of an activity can be viewed as a non-linear manifold embedded in a higher-dimensional space. However, in this case, MLP is not a proper decomposition for activity manifold since it may disorder the geometric structure of action sequence.

\subsection{Our Contributions}
In this paper, we propose a new human activity recognition approach based on the manifold representation of 3D joint locations by integrating the advantages of the temporal relation modeling with the manifold embedding. Rather than modeling the dynamical patterns of the sequence explicitly, manifold learning methods preserve the local geometric properties of activity sequence by embedding it into a low-dimensional space. In this way, human activity is denoted as a series of ordered postures residing on a manifold embedded in a high dimensional space. To construct the sequence of meaningful low-dimensional structures on an activity manifold, we design an efficient division method to decompose an action sequence into the ordered CMLPs based on the nonlinearity degree. Different from the division method proposed in \cite{yao2018human} which divides the action sequence into two sub-sequences each time, our division algorithm is more flexible in that an action sequence can be divided into more than two sub-sequences according to a predefined threshold.

The CMLP corresponding to an action snippet is regarded as a local maximal linear subspace. Motivated by the Cov3DJ proposed in \cite{hussein2013human}, we combine the major posture of action snippet with the main direction of evolution to represent the local maximal linear subspace. In particular, the major posture and main direction are computed with the mean of joints locations and the first principal component of the corresponding covariance matrix, respectively. Based on the intuition that a reasonable distance measure between actions snippets should take both the major posture distance (MPD) and the main direction  distance (MDD) between action snippets into consideration, we define the activity manifold-manifold distance (AMMD) as the pairwise matching of adjacent action snippets in the reference and the test activity manifolds to preserve the local order of action snippets. Our approach is evaluated on two popular benchmarks datasets, KARD dataset \cite{gaglio2015human} and Cornell Activity Dataset (CAD-60) \cite{sung2012unstructured}. Experimental results show the effectiveness and competitiveness of the proposed approach in comparison with the state-of-the-art methods.

In summary, the main contributions of this paper include three aspects:
\begin{itemize}
\item We design an efficient division method to decompose an activity manifold into ordered continuous maximal linear patches (CMLPs) with $k$ sequential neighbors graph.
\item A reasonable distance measure between CMLPs that takes into account both the major posture and the main direction of an action snippet is defined.
\item Based on the distance between CMLPs, an activity manifold-manifold Distance (AMMD) that incorporates the sequential property of action snippets is proposed to discriminate different activities.
\end{itemize}

\section{The Proposed Approach}
\label{sec:recognition}
This section presents the proposed approach for human activity recognition. We first describe the algorithm for the construction of continuous maximal linear patch (CMLP), which decomposes an activity manifold into a sequence of CMLPs viewed as action snippets. Next, we represent CMLP with major posture and main direction, and propose the definition of the distance measure between CMLPs based on this representation. Finally, the activity manifold-manifold distance (AMMD) is computed to discriminate the different activities.

\begin{figure*}[!t]
\centering
\includegraphics[width=\textwidth]{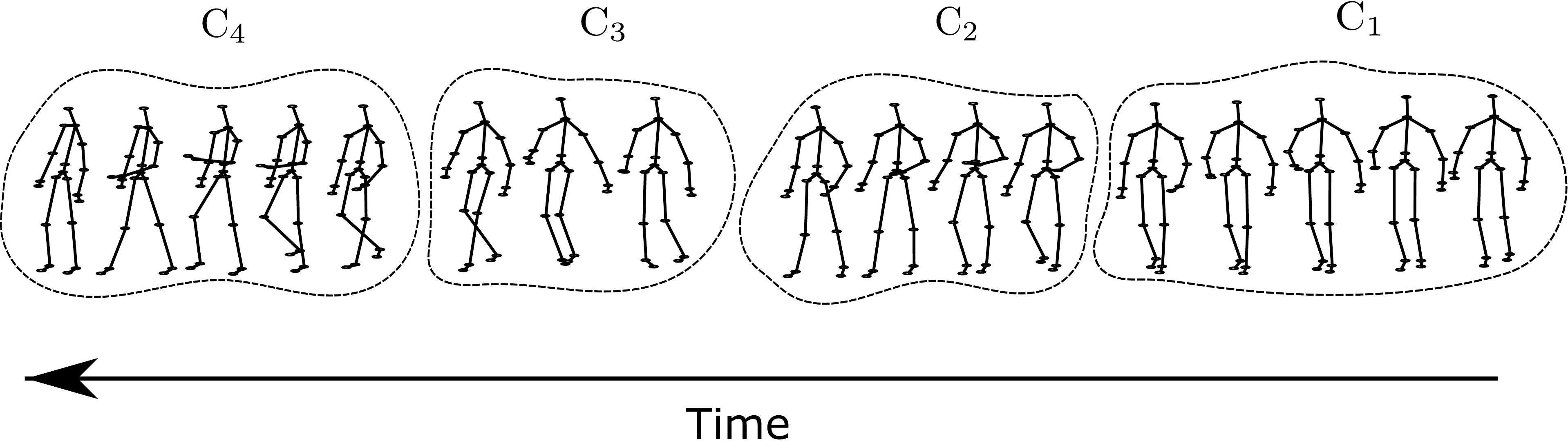}
\caption{The illustration of continuous maximal linear patch (CMLP) construction. The division algorithm based on nonlinear score divides a sequence of postures into several CMLPs. The CMLPs in dotted local patches indicate action snippets.}
\label{fig:cmlp}
\end{figure*}

\subsection{Continuous Maximal Linear Patch}
Local linear models on a manifold are linear patches, whose linear perturbation is characterized by the deviation of the geodesic distances from the Euclidean distances between points. Here the Euclidean distance and the geodesic distance are computed with $l_{2}$-norm and the Dijkstra's algorithm, respectively. The Dijkstra's algorithm is based on the $k$ nearest neighbors graph in which each vertex is connected to its nearest $k$ vertices with the Euclidean metric.

We extend the previous MLP \cite{wang2012manifold} to a new concept termed continuous maximal linear patch (CMLP). The aim of the construction algorithm is to guarantee that each CMLP only contains meaningful successive postures so that it can be regarded as an action snippet. In view of a rational hypothesis that adjacent postures would close to each other in Euclidean metric, we define a $k$ sequential neighbors graph to compute the geodesic distance between postures in action sequence as follows:

\begin{definition}
\textbf{$k$ sequential neighbors graph:} A graph in which each vertex are connected to its previous and next $k$ vertices in temporal order.
\end{definition}

Formally, a human activity is a sequence of human postures $\mathbf{P}=\{\mathbf{p}_1,\mathbf{p}_2,\cdots,\mathbf{p}_F\}$, where $\mathbf{p}_f \in \mathbb{R}^{D}$ is a $D$-dimensional column vector ($D=3\times J$ is 3 coordinates of $J$ human joints), and $F$ is the number of postures. Assume these postures lie on a low-dimensional manifold $\mathcal{M}$ composed of several subspaces, we aim to construct a sequence of CMLPs $\textmd{C}_{i}$ from $\textmd{P}$,
\begin{equation*}
\begin{split}
&\textmd{P}=\{\textmd{C}_1,\textmd{C}_2,\cdots,\textmd{C}_m\}, \\
&\textmd{C}_{i}|_{i=1}^{m}=[\mathbf{p}^i_1,\mathbf{p}^i_2,\cdots,\mathbf{p}^i_{F_i}], \quad \sum_{i=1}^{m} F_i = F,
\end{split}
\end{equation*}
where $m$ is the total number of CMLPs and each action snippet $\textmd{C}_{i}$ contains $F_{i}$ postures.

An efficient division method based on nonlinear score is proposed to construct CMLP. In particular, the current action snippet only contains the first posture, and we include the next posture to current action snippet until the nonlinear score of current action snippet exceeds a defined threshold $\delta$. The next action snippet initialized with empty set continues this process. An illustration of constructed CMLPs is presented in Figure \ref{fig:cmlp}. The nonlinearity score $\beta_i$ to measure the CMLP nonlinearity degree is defined as in \cite{wang2012manifold},
\begin{equation}
\label{eq:nonlinearity}
\beta_i = \frac{1}{F_i^{2}} \sum_{t \in \textmd{C}_{i}} \sum_{s \in \textmd{C}_{i}} r(\mathbf{p}_t,\mathbf{p}_s),
\end{equation}
where $r(\mathbf{p}_t,\mathbf{p}_s)=d_{G}(\mathbf{p}_t,\mathbf{p}_s) / d_{E}(\mathbf{p}_t,\mathbf{p}_s)$ is the ratio of the geodesic distance $d_{G}$ and the Euclidean distance $d_{E}$ computed by $k$ sequential neighbors graph. We average the ratios between each pair of postures $\mathbf{p}_t$ and $\mathbf{p}_s$ in $\textmd{C}_{i}$ to obtain a robust measurement of nonlinearity degree, and the computation of $\beta_i$ can be efficiently carried out.

\begin{algorithm}[!t]
\caption{Construction of Continuous Maximal Linear Patch (CMLP).}
\label{alg:Framwork}
\begin{algorithmic}[1]
\STATE{\textbf{Input:}} \\
~~ A activity sequence $\mathbf{P} = \{\mathbf{p}_1,\mathbf{p}_2,\cdots,\mathbf{p}_F\}$; \\
~~ The nonlinearity degree threshold $\delta$; \\
~~ The number of sequential neighbors $k$.
\STATE{\textbf{Output:}} \\
~~ Local linear model sequences $\textmd{C}_{i}|_{i=1}^{m}$.
\STATE Initialization: \\
~~ $\textmd{C} = \varnothing$, $i = 1$, $\textmd{C}_{i} = \varnothing$, $\beta^{(i)} = 0$, $f=1$; \\
~~ Euclidean distance matrix $D_{E} = \varnothing$; \\
~~ Geodesic distance matrix $D_{G} = \varnothing$; \\
~~ Distance ratio matrix $R = \varnothing$;
\WHILE {$\mathbf{P} \neq \varnothing$}
\STATE Update $\textmd{C}_{i} = \textmd{C}_{i} \cup \mathbf{p}_f$, $\mathbf{P} = \mathbf{P} - \{\mathbf{p}_f\}$;
\STATE Expand $D_E$, $D_G$, $R$ to include $\mathbf{p}_f$;
\STATE Compute the nonlinearity score $\beta^{(i)}$ with Eq. \eqref{eq:nonlinearity};
\IF {$\beta^{(i)} > \delta$}
\STATE Update $\textmd{C} = \textmd{C} \cup \textmd{C}_{i}$, $i = i + 1$;
\STATE Reset $\textmd{C}^{(i)}=\varnothing$, $\beta^{(i)} = 0$,
\STATE \qquad $D_E = \varnothing$, $D_G = \varnothing$, $R = \varnothing$;
\ENDIF
\STATE Update $f = f + 1$;
\ENDWHILE
\IF {$\textmd{C}_{i} \neq \varnothing $}
\STATE Update $\textmd{C} = \textmd{C} \cup \textmd{C}_{i}$;
\ENDIF
\RETURN $\textmd{C}$;
\end{algorithmic}
\end{algorithm}

The improved CMLP not only inherits the ability of MLP to span a maximal linear patch, but also holds the intrinsic structure of successive postures which imply the evolution of corresponding human action snippet. A nonlinearity degree threshold $\delta$ is utilized to control the trade-off between the accuracy of representation and the range of a CMLP. Specifically, a smaller $\delta$ leads to a better accurate representation but a shorter range, and vice versa. Obviously, to make the algorithm applicable, $\delta$ is supposed to be specified to a value larger than $1$ to construct meaningful CMLP sequence. The algorithm of construction of CMLP is summarized in Algorithm \ref{alg:Framwork}. The index $f$ and $i$ indicate the current posture and the current CMLP, respectively. After the initialization of the distance matrix and the distance ratio matrix, we include current posture $\mathbf{p}_f$ into current CMLP and compute the nonlinear score of current CMLP. If the nonlinear score is greater than threshold $\delta$, we obtain the first CMLP and reset distance matrix and distance ratio matrix to the initial values. Otherwise, the index of the current posture is assigned to the next posture. These procedures continue until the entire sequence is divided into several CMLPs.

\subsection{Distance Measure between CMLPs}
An activity manifold is decomposed into ordered CMLPs, and each CMLP can be regarded as a linear patch spanned by the continuous postures. We represent a linear patch with its center and the first principal component of the covariance matrix, which specify the major posture and main direction of the evolution of an action snippet, respectively.

For a CMLP $\textmd{C}_{i}$ denoted by a sequence of postures $\left[\mathbf{p}_1^i,\mathbf{p}_2^{i},\cdots,\mathbf{p}^i_{F_i}\right]$, the major posture is averaged on all postures in this CMLP,
\begin{equation}
\mathbf{u}_i=\frac{1}{F_i} \sum_{f=1}^{F_i}{\mathbf{p}^i_f},
\end{equation}
where $\mathbf{p}^i_f$ is the $f$-th posture of the CMLP $\textmd{C}_{i}$. The sample covariance matrix can be obtained with the formula,
\begin{equation}
\mathbf{\Sigma}_i = \frac{1}{F_i} \sum_{f=1}^{F_i}{(\mathbf{p}^i_f-\mathbf{u}_i)(\mathbf{p}^i_f-\mathbf{u}_i)^\top}.
\end{equation}
By performing eigen-decomposition on the symmetric matrix $\mathbf{\Sigma}_i$, the covariance matrix can be factorized as
\begin{equation}
\mathbf{Q}^{-1}\mathbf{\Lambda} \mathbf{Q}=\mathbf{\Sigma}_i,
\end{equation}
where the diagonal matrix $\mathbf{\Lambda}$ contains the real eigenvalues of $\mathbf{\Sigma}_i$ on its diagonal elements, and $\mathbf{Q}$ is the orthogonal matrix whose columns are the eigenvectors of $\mathbf{\Sigma}_i$ and corresponds to the eigenvalues in $\mathbf{\Lambda}$. The eigenvector that is associated with the largest eigenvalue of $\mathbf{\Sigma}_i$ is denoted by $\mathbf{v}_i$.

For distance measure between two subspaces, the commonly used method is principal angles \cite{bjorck1973numerical}, which is defined as the minimal angles between any two vectors of the subspaces. In particular, let $\mathcal{S}_1$ and $\mathcal{S}_2$ be subspaces of $\mathbb{R}^D$ with dimensions $d_1$ and $d_2$ respectively, and $d=\min(d_1,d_2)$. The $j$-th principal angle $0\leq\theta_j\leq\pi/2\ (j\in\{1,\cdots,d\})$ between $\mathcal{S}_1$ and $\mathcal{S}_2$ are defined recursively as follows,
\begin{equation}
\begin{aligned}
&\cos(\theta_j)=\max\limits_{\mathbf{x}_j\in\mathcal{S}_1}\max\limits_{\mathbf{y}_j\in\mathcal{S}_2} \mathbf{x}_j^\top\mathbf{y}_j, \\
\text{s.t.} \ & \Vert\mathbf{x}_j\Vert=\Vert\mathbf{y}_j\Vert=1, \ \mathbf{x}_j^\top\mathbf{x}_i=\mathbf{y}_j^\top\mathbf{y}_i=0, \\
& \text{where}\ i=1,\cdots,j-1.
\end{aligned}
\end{equation}
The vector pairs $(\mathbf{x}_j,\mathbf{y}_j)$ are called the $j$-th principal vectors. Denote the orthonormal bases of $\mathcal{S}_1$ and $\mathcal{S}_2$ with $\mathbf{S}_1$ and $\mathbf{S}_2$, respectively, the principal angles can be computed straightforward based on the singular value decomposition of $\mathbf{S}_1^\top\mathbf{S}_2$. Concretely, the cosine of the $j$-th principle angle is the $j$-th singular value of $\mathbf{S}_1^\top\mathbf{S}_2$.

Various subspace distance definitions have been proposed based on principal angles. For example, max correlation and min correlation are defined using the smallest and largest principal angles, respectively, while \cite{edelman1999geometry} employed all principal angles in their subspace distance. However, these definitions fail to reflect the difference of subspace positions since principal angles only characterize the difference in direction variation. To derive a better distance measure between CMLPs, we take both the subspace position and direction variation into consideration to measure the main posture distance (MPD) and main direction distance (MDD) between the corresponding action snippets, respectively. The MPD between two CMLPs $\textmd{C}_{i}$ and $\textmd{C}_{j}$ is related to the cosine similarity of $\mathbf{u}_i$ and $\mathbf{u}_j$,
\begin{equation}
\begin{aligned}
\label{eq:MPD}
d_{P}(\textmd{C}_{i},\textmd{C}_{j}) &= (1 - \cos^{2}\alpha)^{1/2} = \sin\alpha, \\
\cos\alpha &=\frac{\mathbf{u}_{i}^\top\mathbf{u}_{j}}{\left(||\mathbf{u}_{i}||\cdot||\mathbf{u}_{j}||\right)}.
\end{aligned}
\end{equation}
In contrast to previous work that assigns weights to each eigenvector, in our case, the MDD is simply defined as the sine distance between first eigenvectors $\mathbf{v}_{i}$ and $\mathbf{v}_{j}$ of two CMLPs $\textmd{C}_{i}$ and $\textmd{C}_{j}$,
\begin{equation}
\begin{aligned}
\label{eq:MDD}
d_{D}({\textmd{C}_{i},\textmd{C}_{j}}) &= (1 - \cos^{2}\beta)^{1/2} = \sin\beta, \\
\cos\beta &=\frac{\mathbf{v}_{i}^\top\mathbf{v}_{j}}{\left(||\mathbf{v}_{i}||\cdot||\mathbf{v}_{j}||\right)}.
\end{aligned}
\end{equation}

The employment of sine distance on both MPD and MDD leads to our distance definition between CMLPs,
\begin{equation}
\begin{aligned}
d_c(\textmd{C}_{i},\textmd{C}_{j}) &= \bigg(d_{P}^2(\textmd{C}_{i},\textmd{C}_{j}) + d_{D}^2({\textmd{C}_{i},\textmd{C}_{j}})\bigg)^{1/2} \\
&= (\sin^2\alpha + \sin^2\beta)^{1/2} \\
&= (2 - \cos^2\alpha - \cos^2\beta)^{1/2}.
\end{aligned}
\end{equation}
This distance is then used as the basis for the following distance measure between action manifolds.

\begin{figure*}[!t]
\centering
\includegraphics[width=0.8\textwidth]{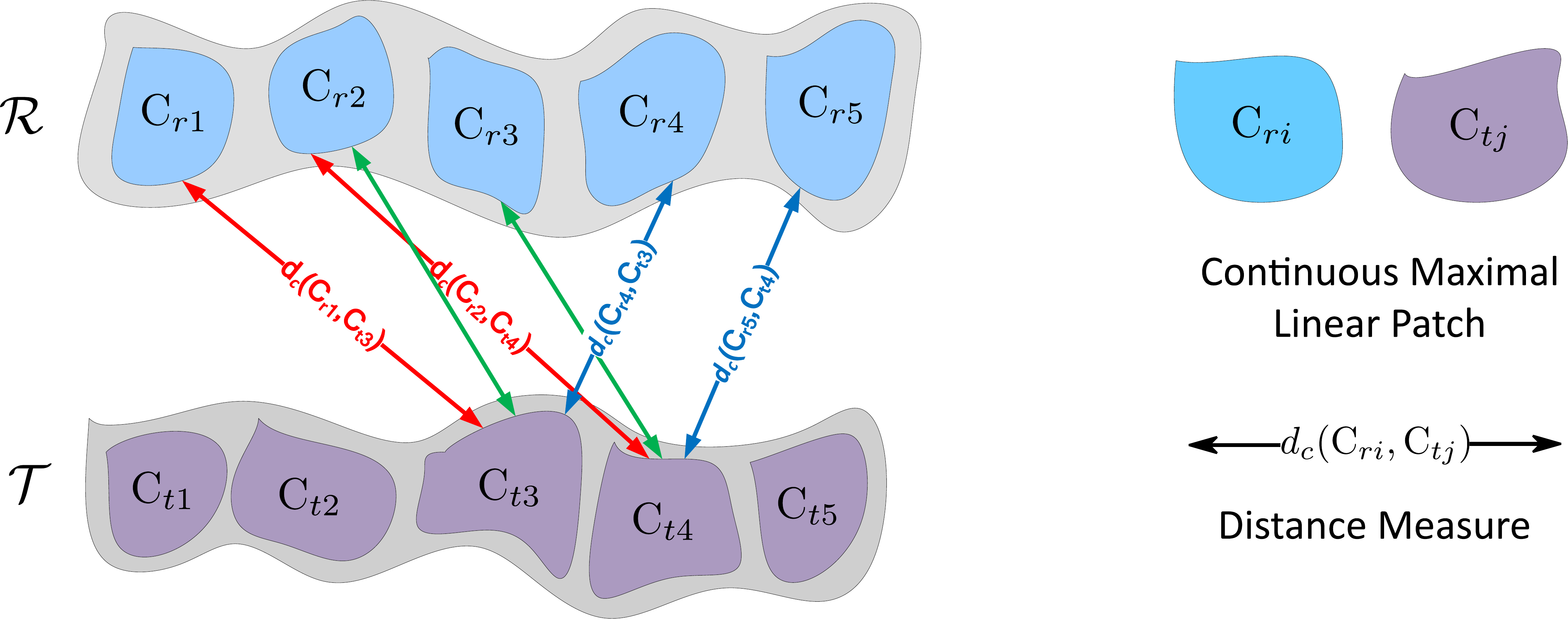}
\caption{Illustration of the proposed pairwise distance between Continuous Maximal Linear Patches (CMLPs). $\mathcal{R}$ and $\mathcal{T}$ denote the reference and test action manifold, respectively. $\mathrm{C}_{ri}$ and $\mathrm{C}_{tj}$ are the $i$-th and $j$-th CMLP in the reference and the test action manifold. To preserve the local order of action snippets, we match each pair of adjacent two CMLPs between the reference action manifold and the test action manifold. The distance measure is indicated by the double-sided arrow in different colors.}
\label{fig:PPDistance}
\end{figure*}

\subsection{Activity Manifold-Manifold Distance}
Given the reference and test activity manifold denoted as $\mathcal{R} = \{\textmd{C}_{r1}, \textmd{C}_{r2},\cdots,\textmd{C}_{rm}\}$ and $\mathcal{T} = \{\textmd{C}_{t1}, \textmd{C}_{t2},\cdots,\textmd{C}_{tn}\}$, respectively, where $\textmd{C}_{ri}$ and $\textmd{C}_{tj}$ are CMLPs, we aim to measure Activity Manifold-Manifold Distance (AMMD) based on the distance between CMLPs. An intuitive definition of the manifold to manifold distance is proposed in \cite{wang2012manifold},
\begin{equation}
\begin{aligned}
d(\mathcal{R},\mathcal{T})&=\sum_{i=1}^m\sum_{j=1}^n w_{ij}d_c(\textmd{C}_{ri},\textmd{C}_{tj}), \\
\text{s.t.}&\ \sum_{i=1}^m\sum_{j=1}^n w_{ij}=1,\ w_{ij}\geq 0.
\end{aligned}
\end{equation}
This definition integrates all pairwise subspace to subspace distance and is a many-to-many matching problem. The difficulty is how to determine the weight $w_{ij}$ between subspaces $\textmd{C}_{ri}$ and $\textmd{C}_{tj}$. Although earth mover's distance ($1$-st Wasserstein distance) \cite{rubner2000earth} can be employed compute $w_{ij}$, its computational complexity is too high. In practice, all weights are set as an equal constant value $\frac{1}{m+n}$. 

In the scenario of face recognition with image set (FRIS) \cite{wang2012manifold}, the authors believed that the closet subspace pair deserves the most emphasis and defined the manifold to manifold distance as the distance of closest subspace pair from these two manifolds as follows,
\begin{equation}
\begin{aligned}
d(\mathcal{R},\mathcal{T})=&\min\limits_{\textmd{C}_{ri}\in\mathcal{R}}d(\textmd{C}_{ri},\mathcal{T}) \\
=&\min\limits_{\textmd{C}_{ri}\in\mathcal{R}}\min\limits_{\textmd{C}_{tj}\in\mathcal{T}}d_c(\textmd{C}_{ri},\textmd{C}_{tj}).
\end{aligned}
\end{equation}
It is easy to find out that the weight of the closet pair is set to $1$ and all the other weights are set to $0$ in this case. The best-suited subspaces distance is one of the most appropriate manifold-manifold distances for FRIS problem. However, it cannot be applied to our activity recognition problem since this distance ignores the temporal relationship between actions snippets. To preserve the local order of action snippets in distance definition, we propose to match the pairwise adjacent two CMLPs from the test manifold to the reference manifold and obtain the following distance,
\begin{equation}
\begin{aligned}
&d(\mathcal{R},\mathcal{T}) \\
=&\sum_{j=1}^{m-1} \min_{i \in [1,n)}\left[d_c(\textmd{C}_{ri},\textmd{C}_{tj})+d_c(\textmd{C}_{r,i+1},\textmd{C}_{t,j+1})\right].
\end{aligned}
\end{equation}
As illustrated in Figure \ref{fig:PPDistance}, for each CMLP pair extracted in test manifold $\mathcal{T}$, we find the most similar pair from reference action manifold $\mathcal{R}$, and the sum of all pairwise distances amounts to the AMMD. Afterward, the unknown activity is assigned to the class that has the closest AMMD over all reference action classes,
\begin{equation}
\text{label}\quad l=\argmin\limits_{c}\{d(\mathcal{R}_{c}, \mathcal{T})\},
\end{equation}
where $d(\mathcal{R}_c,\mathcal{T})$ is the distance between the $c$-th class reference action manifold and the test action manifold.

\section{Experiments}
\label{sec:experiment}
We study the performance of our approach on two popular benchmarks, KARD dataset \cite{gaglio2015human} and Cornell Activity Dataset (CAD-60) \cite{sung2012unstructured}. Both of them records 15 joint locations for the participated subjects. In all experiments, the hyperparameters, the number of linked sequential neighbor and the nonlinearity degree threshold, are selected with cross-validation.

\subsection{KARD Dataset}

\begin{table}[tbp]
\caption{Subset segmentation of KARD dataset. The ten Actions are indicated in bold font.}
\label{tab:KARDGroup}
\renewcommand{\arraystretch}{1.2}
\centering
{\small
\begin{tabular}{p{2.65cm}|p{2cm}|p{2.65cm}}
\hline \hline
Subset 1 & Subset 2 & Subset 3 \\ \hline
Horizontal arm wave & High arm wave & Draw tick \\
Two-hand wave & Side kick & \textbf{Drink} \\
Bend & \textbf{Catch cap} & \textbf{\small{Sit down}} \\
\textbf{Phone call} & Draw tick & \textbf{Phone call} \\
\textbf{Stand up} & Hand clap & \textbf{Take umbrella} \\
Forward kick & Forward kick & \textbf{Toss paper} \\
Draw X & Bend & High throw \\
\textbf{Walk} & \textbf{Sit down} & Horizontal arm wave \\
\hline \hline
\end{tabular}}
\end{table}

The KARD dataset contains 18 activities collected by Gaglio et al. \cite{gaglio2015human}. These activities include ten gestures and eight actions, and are grouped into three subsets as listed in Table \ref{tab:KARDGroup}. The obtained sequences are collected on 10 different subjects that perform each activity 3 times, thus, there are 540 skeleton sequences in this dataset. According to the previous work \cite{gaglio2015human}, KARD dataset is split under three different setups and two modalities in the experiment. Specifically, the three experiments setups A, B, and C utilize one-third, two-thirds, and half of the samples for training, respectively, and the rest for testing. The activities constituting the dataset are split into the five groups: Gestures, Actions, Activity Set 1, 2, and 3 (three subsets). From subset 1 to 3, the activities become increasingly difficult to recognize due to the increase of similarity between activities. Note that Actions are more complex than Gestures.

\begin{table*}[t]
\caption{Accuracies on the KARD dataset under three different experimental setups of five different splittings.}
\label{tab:KARDResultsAll}
\centering
\resizebox{\textwidth}{!}{
\renewcommand{\arraystretch}{1.2}
\begin{tabular}{|c|ccc|ccc|ccc|ccc|ccc|}
\hline
\multirow{2}{*}{Methods}& \multicolumn{3}{|c|}{Subset 1} & \multicolumn{3}{c|}{Subset 2} & \multicolumn{3}{c|}{Subset 3} & \multicolumn{3}{c|}{Gestures} & \multicolumn{3}{c|}{Actions} \\ \cline{2-16}
 & \multicolumn{1}{|c}{A} & B & C & A & B & C & A & B & C & A & B & C & A & B & C \\ \hline
\cite{gaglio2015human} & \multicolumn{1}{|c}{95.1} & 99.1 & 93.0 & 89.9 & 94.9 & 90.1 & 84.2 & 89.5 & 81.7 & 86.5 & 93.0 & 86.7 & 92.5 & 95.0 & 90.1 \\
\hline
\cite{cippitelli2016human} & \multicolumn{1}{|c}{98.0} & 99.0 & 97.7 & 99.8 & 100 & 99.6 & 91.6 & 95.8 & 93.3 & 89.9 & 95.9 & 93.7 & \textbf{99.0} & \textbf{99.9} & \textbf{99.1} \\ \hline
The Proposed Approach & \multicolumn{1}{|c}{\textbf{100}} & \textbf{100} & \textbf{100} & \textbf{99.9}& \textbf{100} & \textbf{99.8} & \textbf{97.6} & \textbf{98.0} & \textbf{96.8} & \textbf{99.6} & \textbf{99.8} & \textbf{99.9}& 97.6 & 98.1 & 96.9 \\ \hline
\end{tabular}
}
\end{table*}

\begin{table}[t]
\caption{Accuracies on the KARD dataset under the ``new-person" setting.}
\label{tab:KARDResultsNewPerson}
\centering
\renewcommand{\arraystretch}{1.2}
{\small
\begin{tabular}{|c|c|}
\hline
Methods & Accuracy \% \\ \hline
\cite{gaglio2015human} & 84.8\\ \hline
\cite{cippitelli2016human} & 95.1\\ \hline
The Proposed Approach & \textbf{99.3}\\ \hline
\end{tabular}
}
\end{table}

\begin{figure}[t]
\centering
\includegraphics[width=\linewidth]{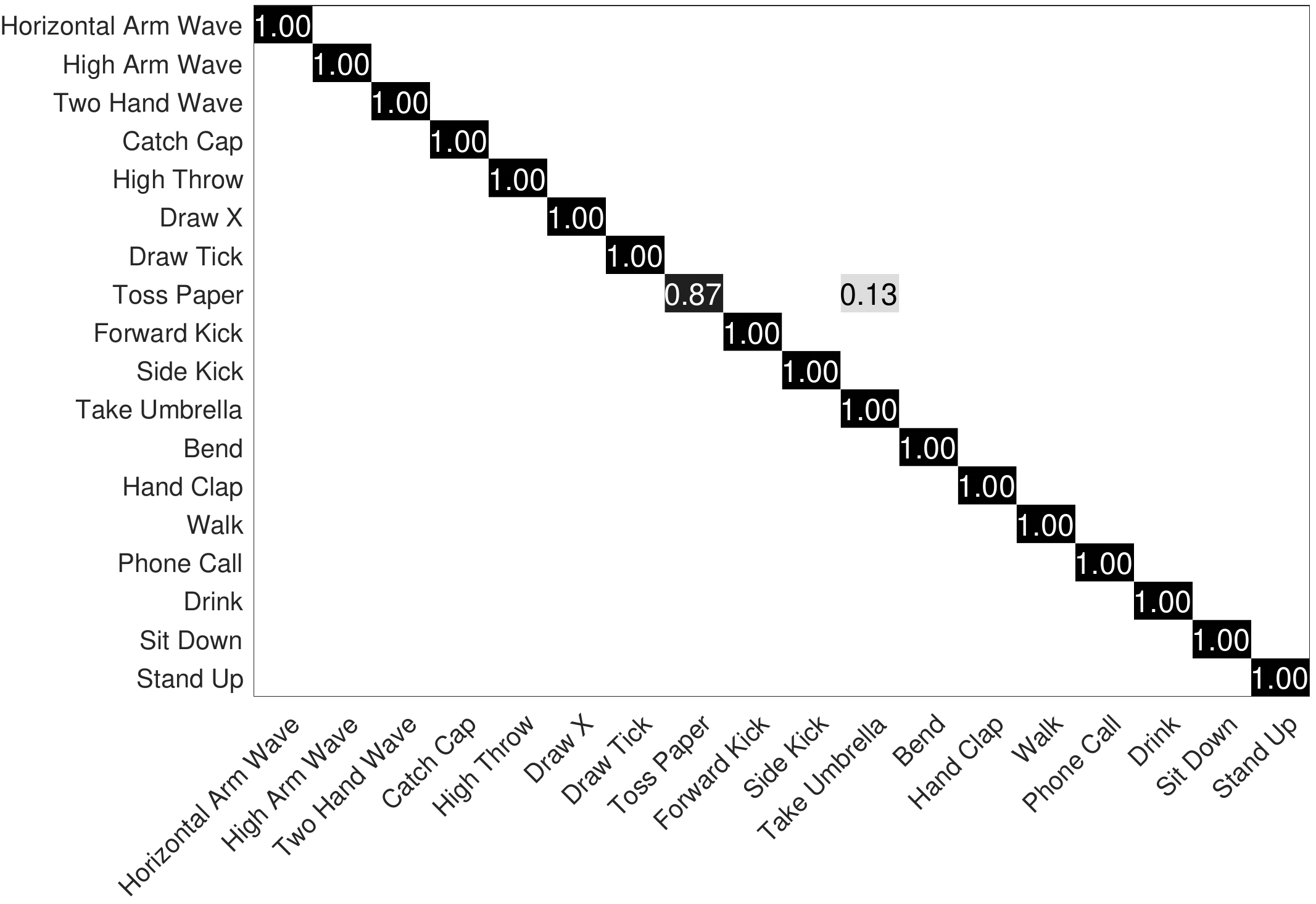}
\caption{Confusion matrix on the KARD dataset under the ``new-person'' setting.}
\label{fig:KARDConfusionMatrix}
\end{figure}

All results on this dataset are obtained with the parameter setting: $k = 5$, $\delta = 1.04$. In consideration of the randomness existing in the dataset splitting procedure, we run each experimental setup 10 times and present the mean performance in Table \ref{tab:KARDResultsAll}. The proposed approach outperforms all other methods on four out of five subsets under all experimental setups but narrowly lost to the method in \cite{cippitelli2016human} on the Actions subset. The reason is that the CMLP representation is a linear descriptor, which might fail to capture some nonlinear features of complex activities and is unable to discriminate the subtle difference between the similar activities as a result.

In addition, we perform the experiment in the ``new-person" scenario, i.e., a leave-one-subject-out setting. The experimental setting is in line with that in \cite{cippitelli2016human}. Table \ref{tab:KARDResultsNewPerson} presents the results of the proposed approach compared with the state-of-the-arts. It can be observed that our approach achieves the best result with an accuracy of $99.3\%$, which exceeds the second best result by $4.2\%$. Figure \ref{fig:KARDConfusionMatrix} illustrates the confusion matrix, which shows that the proposed approach classifies all activities correctly with only slight confusion between activities $toss$ $paper$ and $take$ $umbrella$. The reason is that the representations with 3D joint locations are almost the same in these two activities, and the proposed approach is prone to confuse the activities based on the limited information obtained from the linear descriptor. This confusion directly degrades the performance of the experimental setup  ``Actions" in Table \ref{tab:KARDResultsAll}. We believe that it would be sensible to explore the addition of RGB or depth image information in our future work. In summary, this newly proposed approach achieved impressive performance on the above human activity recognition tasks in our current experimental setting.

\subsection{Cornell Activity Dataset}
\begin{table}[t]
\caption{Accuracies on the CAD-60 dataset under the ``cross-person'' setting.}
\label{tab:CAD-60ResultsNewPerson}
\centering
\renewcommand{\arraystretch}{1.2}
{\small
\begin{tabular}{|c|c|}
\hline
Methods & Accuracy \% \\ \hline
\cite{wang2014learning}  & 74.7 \\ \hline
\cite{koppula2013learning}  & 80.8\\ \hline
\cite{hu2015jointly} & 84.1 \\ \hline
\cite{cippitelli2016human} & 93.9 \\ \hline
The Proposed Approach & \textbf{99.6} \\ \hline
\end{tabular}
}
\end{table}

\begin{figure}[t]
\centering
\includegraphics[width=\linewidth]{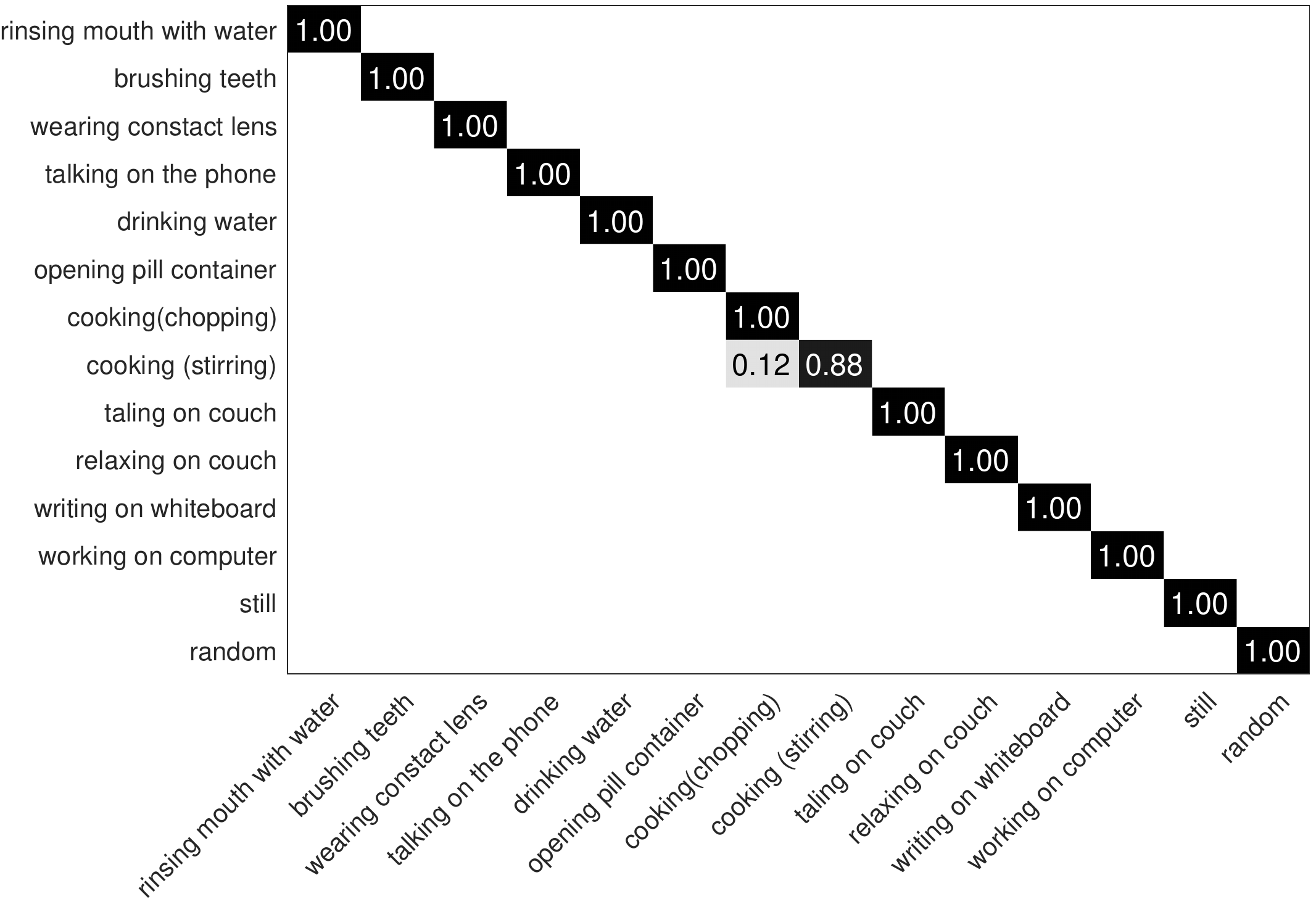}
\caption{Confusion matrix on the CAD-60 dataset under the ``cross-person'' setting.}
\label{fig:CAD60ConfusionMatrix}
\end{figure}

Cornell Activity Dataset 60 (CAD-60) \cite{sung2012unstructured} is a human activity dataset comprising of twelve unique activities. Four different human subjects (one is left-handed and others are right-handed, two males and two females) are asked to perform three or four common activities in five different environments, including bathroom, bedroom, kitchen, living room and office. The experimental setting of leave-one-person-out cross-validation is adopted as in \cite{wang2014learning} that the person in the training would not appear in the testing for each environment. To eliminate the influence from the left-handed subject, if the $y$-coordinate of the right hand is smaller than the left hand, we interchange the $y$-coordinate of left and right hands, ankles and shoulders, to transform skeleton positions of the left-handed persons to those of the right-handed ones.

Here, the number of sequential neighbors and nonlinearity degree threshold are set as $k=1$ and $\delta = 1.2$, respectively. The recognition performance is shown in Table \ref{tab:CAD-60ResultsNewPerson} by averaging the accuracies on all possible splits (totally 20). The proposed approach achieves an accuracy of $99.6\%$, which outperforms the results of the comparative methods. Figure \ref{fig:CAD60ConfusionMatrix} shows the confusion matrix of the performance obtained by our proposed approach. It can be observed that the proposed approach classifies all the actions correctly except minor confusion on two $cooking$ actions: $stirring$ and  $chopping$, which is probably caused by the inaccurate human skeletons information. In conclusion, the appealing recognition result demonstrates that our approach can effectively capture the evolution of human activities only based on human 3D joints locations.

\begin{figure}[!t]
\centering
\includegraphics[width=\linewidth]{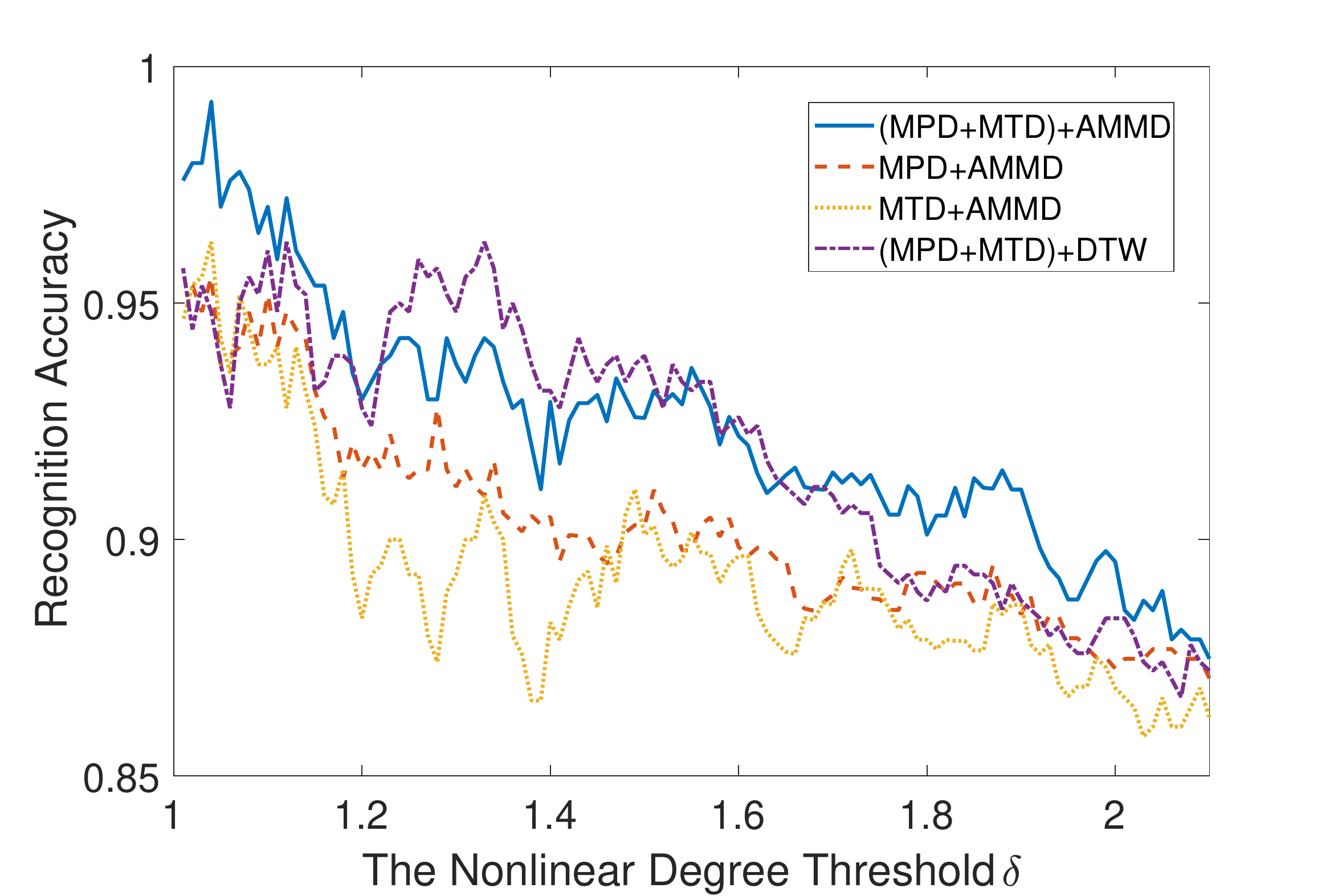}
\caption{The recognition accuracies of different distance measurements on the KARD dataset with respect to the threshold $\delta$ under the ``new person" setting.}
\label{fig:KARDAnalysis}
\end{figure}

\subsection{Parameter Analysis} 
There are two key parameters in our approach: the nonlinearity degree threshold $\delta$ and the number of linked sequential neighbors $k$. The nonlinearity degree threshold determines the granularity of continuous linear maximal patches, while the number of linked sequential neighbors quantifies the topology preservation in the computation of geodesic distance. To evaluate the sensitivity of the proposed approach with respect to these two parameters, we conduct experiments with different parameter values under the leave-one-subject-out setting on the KARD dataset.

\begin{figure}[!t]
\centering
\includegraphics[width=\linewidth]{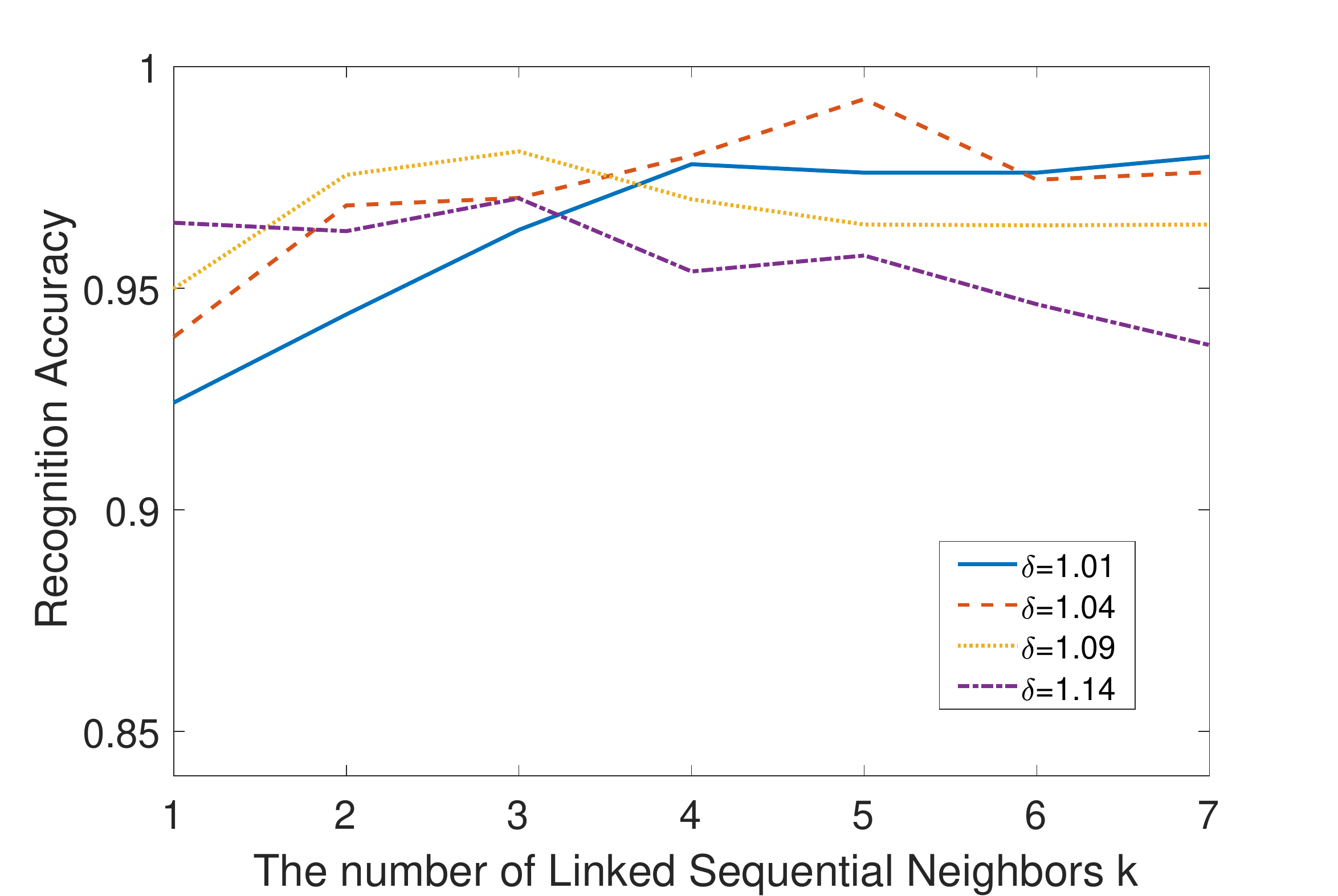}
\caption{The recognition accuracy of the proposed approach on KARD dataset with respect to the number of sequential neighbors $k$ under the ``new-person" setting.}
\label{fig:KARDAnalysisK}
\end{figure}

We first fix the number of neighbors $k$ and adjust the nonlinearity degree threshold $\delta$ from $1$ to $2.1$ with step size $0.01$. Figure \ref{fig:KARDAnalysis} illustrates the corresponding experimental performance. The relatively small gap between the worst and the best results under each distance measurement validates that the proposed approach is quite robust with respect to the value of $\delta$. Generally speaking, lower $\delta$ could lead to a better performance since a smaller CMLP yields more representative action snippet. Then, we adjust the number of neighbors $k$ from $1$ to $7$ with step size $1$ and fix the nonlinearity degree threshold $\delta = 1.01,1.04,1.09,1.14$, respectively. The obtained result is illustrated in Figure \ref{fig:KARDAnalysisK}, which shows that with the increase of $k$, the recognition accuracy gets increasingly higher when $\delta$ is small. However, while $\delta$ is large, the recognition accuracy shows a decline with the increase of $k$.

Overall, the best recognition accuracy is usually obtained with a small $k$ matched with a large $\theta$ or vice verse. In some sense, two parameters are not totally independent on determining the final performance of our approach, both parameters cooperate with each other to construct the most representative CMLPs.

\subsection{Distance Measure Methods Comparison}

The proposed distance measure includes two parts: main posture distance (MPD) and main direction distance (MDD) between action snippets. Intuitively, MDD is more discriminative than MPD since the evolution of the main direction is more important than the position of subspace in the activity recognition problem. However, the MPD is complementary to MDD to some extent. To demonstrate the strength of the proposed distance measure, we compare the different combination of distance measure between the CMLPs and sequence matching algorithm.

The results are shown in Figure \ref{fig:KARDAnalysis}, in which dynamic time warping (DTW) is a template matching algorithm that calculates an optimal match between two given sequences under some certain restrictions. The curve of MPD is almost above the curve of MDD, which confirms our intuition that the major posture features more important than main tendency feature for recognition. As expected, MPD holds major posture representation and MDD keeps the ability to describe the evolution of action snippet. Thus the combination of these two distance measurements performs the best. In general, the combination MPD+MDD+AMMD obtains the best results in most cases.

\section{Conclusion}
\label{sec:conclusion}
In this paper, we present a novel human activity recognition approach that utilizes a manifold representation of 3D joint locations. Considering that an activity is composed of several compact sub-sequences corresponding to meaningful action snippets, the 3D skeleton sequence is decomposed into ordered continuous maximal linear patches (CMLPs) on the activity manifold. The computation of activity manifold-manifold distance (AMMD) preserves the local order of action snippets and is based on the pairwise distance between CMLPs, which takes into account the major posture and the main direction of action snippets. Experimental results show better performance of our approach in comparison with the state-of-the-art approaches. In practice, there often exists local temporal distortion and periodic patterns in the action sequence. By viewing action snippets as samples from a probability distribution, we attempt to introduce the Wasserstein metric to measure the distance between the action snippets for activity recognition in the future work.

\bibliographystyle{aaai}
\bibliography{reference}

\end{document}